\documentclass[10pt,twocolumn,letterpaper]{article}

\usepackage{cvpr}

\usepackage{multirow}
\usepackage{rotating}
\usepackage{colortbl}
\usepackage{pifont}

\newcommand{\jy}[1]{{\color{black}#1}}
\definecolor{cvprblue}{rgb}{0.21,0.49,0.74}
\usepackage[pagebackref,breaklinks,colorlinks,allcolors=cvprblue]{hyperref}

\def\confName{CVPR}
\def\confYear{2026}

\title{Camouflage-aware Image-Text Retrieval via Expert Collaboration}

\author{
	Yao Jiang$^{1}$
	\hspace{5pt}
    Zhongkuan Mao$^{2}$
	\hspace{5pt}
    Xuan Wu$^{1}$
	\hspace{5pt}
	Keren Fu$^{1,2,*}$
	\hspace{5pt}
	Qijun Zhao$^{1,2}$
	\\
	$^1$College of Computer Science, Sichuan University
        \hspace{5pt} \\
        $^2$National Key Lab of Fundamental Science on Synthetic Vision, Sichuan University
}

\begin{document}
\maketitle
\thispagestyle{empty}
\renewcommand{\thefootnote}{\fnsymbol{footnote}}
\footnotetext[1]{\jy{Corresponding author: Keren Fu \emph{(fkrsuper@scu.edu.cn)}.}}

\begin{abstract}
Camouflaged scene understanding (CSU) has attracted significant attention due to its broad practical implications. However, in this field, robust image-text cross-modal alignment remains under-explored, hindering deeper understanding of camouflaged scenarios and their related applications. To this end, we focus on the typical image-text retrieval task, and formulate a new task dubbed ``camouflage-aware image-text retrieval'' (CA-ITR). We first construct a dedicated camouflage image-text retrieval dataset (CamoIT), comprising $\sim$10.5K samples with multi-granularity textual annotations. Benchmark results conducted on CamoIT reveal the underlying challenges of CA-ITR for existing cutting-edge retrieval techniques, which are mainly caused by objects' camouflage properties as well as those complex image contents. As a solution, we propose a camouflage-expert collaborative network (CECNet), which features a dual-branch visual encoder: one branch captures holistic image representations, while the other incorporates a dedicated model to inject representations of camouflaged objects. A novel confidence-conditioned graph attention (C\textsuperscript{2}GA) mechanism is incorporated to exploit the complementarity across branches. Comparative experiments show that CECNet achieves $\sim$29\% overall CA-ITR accuracy boost, surpassing seven representative retrieval models. The dataset and code will be available at \url{https://github.com/jiangyao-scu/CA-ITR}.
\end{abstract}    
\section{Introduction}
\label{sec:intro}
Camouflaged scene understanding (CSU) is a comprehensive visual analysis task that aims to perceive and understand objects whose visual characteristics are highly similar to those of the surrounding background~\cite{le2019anabranch,fan2022concealed}. As a challenging and rapidly advancing field within computer vision, it has attracted considerable attention owing to its wide-ranging applications in critical domains, including medicine, industry, and agriculture~\cite{fan2022concealed,bi2022rethinkingcod,fan2023advancescod,fan2020pranet,Fan2020InfNet,Liu2023MHNet}.

\begin{figure}[!t]
\centering
\includegraphics[width=0.98\linewidth]{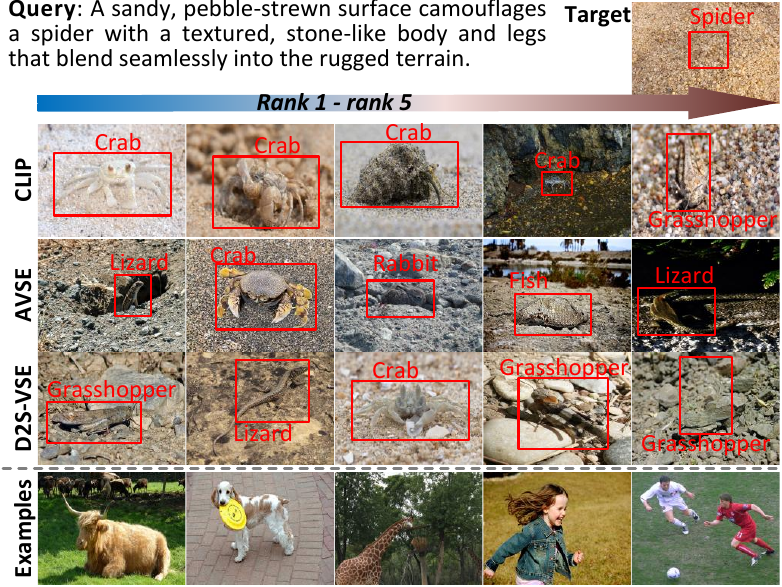}
\vspace{-5pt}
\caption{Qualitative results of three state-of-the-art (SOTA) retrieval methods (\ie, CLIP~\cite{radford2021clip}, AVSE~\cite{liu2025AVSE}, and D2S-VSE~\cite{liu2025D2SVSE}) on CA-ITR. Camouflaged objects in the images are marked with red bounding boxes. Below the dotted line are samples from the general ITR datasets (\ie, MS-COCO~\cite{lin2014coco} and Flickr30K~\cite{young2014flickr}).}
\vspace{-18pt}
\label{fig:qualitative_other}
\end{figure}

The field of CSU has undergone significant methodological evolution in recent years. Early approaches relied solely on visual cues, utilizing sophisticated neural networks to address the challenge of camouflaged object detection (COD)~\cite{wu2023sourcefree,zhong2022codfrequency,zheng2023mffn,fan2022concealed,sun2021contextaware,jia2022Segment,mei2021distractionmining,mei2023distraction,huang2023shrinkage,le2019anabranch}. Subsequent advancements integrated multimodal information to enhance detection performance or reduce reliance on dense annotations~\cite{Li2023ZSCOD,Tang2024MMCPF,Hu2024GenSAM,Zhang2024ACUMEN,Pang2024OVCOS}, thereby improving segmentation accuracy and expanding application scenarios. 
However, the core paradigm of these methods remains largely limited to segmentation tasks, generating pixel-wise masks that inherently lack the rich semantic information required for advanced cognitive tasks.
Recently, emerging studies~\cite{Ruan2025MMCamObj, Zhao2025CvpAgent} have explored multimodal large language models (MLLMs) in camouflaged scenarios, making a step further over the conventional segmentation frameworks. 
Despite such advances, these approaches still predominantly focus on the generative perspective, namely visual question answering and visual grounding, leaving explicit image-text cross-modal alignment largely under-explored.

Robust cross-modal alignment serves as the foundational bedrock for generalized multimodal understanding, underpinning critical capabilities such as visual question answering and detailed image captioning~\cite{Li2024MultimodalAlignmentSurvey,Tadas2019SurveyandTaxonomy}. Failure in such a step would inevitably undermine any downstream task.
As illustrated in Fig.~\ref{fig:qualitative_other}, for image-text retrieval, current state-of-the-art (SOTA) retrieval models frequently mismatch textual descriptions with incorrect visual objects in camouflaged scenes. Meanwhile, the struggles of advanced MLLMs on camouflaged scenes revealed by recent researches~\cite{jiang2024effectiveness,Ruan2025MMCamObj}, can also be likely attributed to this issue, since effective cross-modal understanding usually derives from robust initial alignment~\cite{Li2024MultimodalAlignmentSurvey,Li2021ALBEF}.

In this spirit, we study the camouflage-aware image-text alignment problem via image-text retrieval, which is abbreviated as CA-ITR.
Firstly, we construct a dedicated camouflage image-text retrieval dataset (CamoIT) by 
annotating images from existing COD datasets~\cite{Fan2020COD10K,le2019anabranch,skurowski2018CHAMELEON,Lv2021NC4K} with comprehensive textual descriptions.
CamoIT comprises $\sim$10.5K samples spanning diverse camouflage scenarios. Each image is paired with multi-granularity annotations, including category labels, detailed descriptions of camouflaged objects, and comprehensive image-level captions. 
Leveraging CamoIT, a comprehensive CA-ITR benchmark is conducted by evaluating open-source SOTA retrieval methods/models, 
whose results demonstrate that CA-ITR is not merely a domain transfer task but one that presents several distinct challenges, including the difficulty of perceiving camouflaged objects, complexity of image contents, and necessity for fine-grained understanding.

To study CA-ITR, we further propose a baseline model, namely camouflage-expert collaborative network (CECNet). It features a dual-branch visual encoder: one branch captures holistic image representations, while the other incorporates a dedicated COD model to assist representations of camouflaged objects. A novel confidence-conditioned graph attention (C\textsuperscript{2}GA) mechanism is used to exploit the complementarity across branches.

In a nutshell, our contributions are three-fold:
\begin{itemize}
\item \textbf{Task and dataset.} We formalize the new task of ``camouflage-aware image-text retrieval'' (CA-ITR) and construct CamoIT, which comprises $\sim$10.5K samples across 237 categories. 
CamoIT not only supports CA-ITR but also bridges high-level retrieval with low-level perception. 
To the best of our knowledge, this work is the first attempt in the field to propose and address image-text retrieval in camouflaged scenarios.

\item \textbf{Benchmark and analyses.} We conduct a comprehensive CA-ITR benchmark, 
demonstrating that it is not merely a domain transfer task, but one characterized by unique challenges: perceiving camouflaged objects, handling complex image contents, and achieving fine-grained understanding. 

\item \textbf{New baseline model.} We propose a camouflage-expert collaborative network (CECNet) that incorporates a COD expert to enhance object perception and a novel confidence-conditioned graph attention (C\textsuperscript{2}GA) mechanism to refine feature fusion. Experimental results show that CECNet achieves $\sim$29\% 
overall CA-ITR accuracy boost,
surpassing seven representative retrieval models.
\end{itemize}

\section{Related Work}
\vspace{-1mm}
\label{sec:related}

\begin{figure*}[!ht]
\centering
\includegraphics[width=0.98\linewidth]{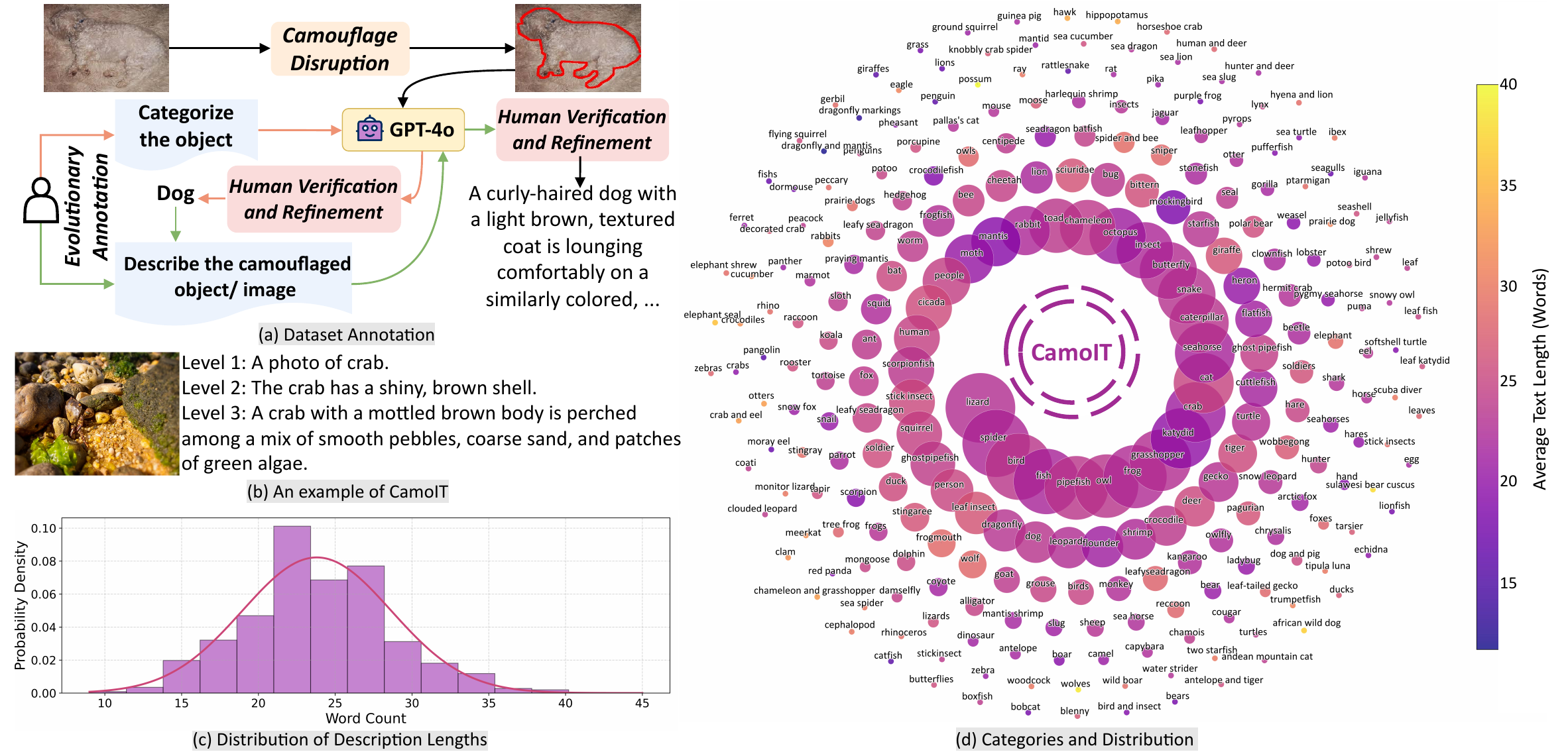}
\vspace{-8pt}
\caption{Data annotation process, an example, and statistical analyses of CamoIT.}
\vspace{-15pt}
\label{fig:DatasetConstruction}
\end{figure*}

\subsection{Image-Text Retrieval}
Image-Text Retrieval (ITR) methods can be divided into global alignment and local alignment. Global alignment methods~\cite{frome2013DeViSE,Faghri2018vsepp,Sarafianos2019Adversarial,li2023Reasoning,zhang2024user,wei2022cfm,fu2023hrem,huang2024cusa,liu2025D2SVSE,liu2025AVSE} match holistic image and text representations, but overlook fine-grained cross-modal correspondences, which limits their retrieval accuracy. In contrast, local alignment methods~\cite{Lee2018Stacked,Wang2019camp,Liu2020Graph,fu2024laps,pan2023CHAN,diao2024dbl} address this limitation to some extent by capturing fine-grained interactions between image regions and text words through mechanisms like cross-attention and graph-based matching. The implementation of these alignment strategies relies on different backbone architectures. 
Several approaches~\cite{wei2022cfm,fu2023hrem,pan2023CHAN} adopt the BUTD framework~\cite{Anderson2017butd}, leveraging object-level visual representations to enhance retrieval. Other methods~\cite{huang2024cusa,fu2024laps,liu2025AVSE,liu2025D2SVSE} utilize learnable image encoders to extract image features. D2S-VSE~\cite{liu2025D2SVSE} further explores a dense text distillation strategy to enhance cross-modal alignment.
Although these methods have made significant progress in ITR, they are mainly trained on data with clearly separated foreground objects, leading to a notable performance drop when handling camouflaged objects that closely resemble the background.

\subsection{Camouflaged Scene Understanding}
Current research in CSU predominantly focuses on camouflaged object detection (COD)~\cite{bi2022rethinkingcod,fan2023advancescod}, with extensions into related tasks such as camouflaged object instance segmentation~\cite{He2024TextpromptCIS,Luo2023CIS}, camouflaged object counting~\cite{Sun2023COC}, and camouflaged object ranking~\cite{Lv2021NC4K}.
These COD methods rely solely on visual cues and employ diverse network architectures, including multi-stream frameworks~\cite{yan2021mirrornet,wu2023sourcefree,zhong2022codfrequency,pang2022zoom,zheng2023mffn,Kajiura2021Improving,Pang2024zoomnext}, bottom-up and top-down integration~\cite{fan2022concealed,sun2021contextaware, jia2022Segment,mei2021distractionmining, mei2023distraction,huang2023shrinkage,Fan2020COD10K}, and auxiliary branches~\cite{le2019anabranch,Lv2021NC4K,li2021uncertaintyaware}.
In recent years, multimodal textual and visual information has been increasingly applied in CSU. Several approaches~\cite{Li2023ZSCOD,Tang2024MMCPF,Hu2024GenSAM} leverage MLLMs to reduce dependence on densely annotated data, utilizing visual reasoning and grounding capabilities to enable camouflaged object segmentation in open-world or zero-shot settings. ACUMEN~\cite{Zhang2024ACUMEN} enhances segmentation by incorporating textual descriptions of camouflaged attributes, while OVCoser~\cite{Pang2024OVCOS} enables open-vocabulary segmentation through target category integration.
These methods are still largely confined to segmentation tasks, producing pixel-wise masks that lack the rich semantic information needed for advanced cognitive tasks.

Recently, several studies~\cite{Ruan2025MMCamObj,Zhao2025CvpAgent} have developed multimodal large models for camouflaged scenarios.
However, these approaches predominantly target generative tasks such as dialogue systems, leaving the core challenge of achieving robust cross-modal alignment in complex camouflaged environments largely unaddressed.

\vspace{-1mm}
\section{Construction of CamoIT}
\vspace{-1mm}
\label{sec:dataset}
Our CamoIT is built upon existing COD datasets (\ie, CHAMELEON~\cite{skurowski2018CHAMELEON}, CAMO~\cite{le2019anabranch}, COD10K~\cite{Fan2020COD10K}, and NC4K~\cite{Lv2021NC4K}), from which we excluded low-quality samples, such as those with unrecognizable categories. As these datasets were designed originally for the segmentation task, they primarily provide mask annotations. To adapt them for the CA-ITR task, we utilized GPT-4o~\cite{hurst2024gpt4o} to support data annotation, enabling the efficient generation of comprehensive textual descriptions.
Our data annotation process consists of three components: camouflage disruption, evolutionary annotation protocol, and manual verification and refinement, as shown in Fig.~\ref{fig:DatasetConstruction} (a). 
The construction adheres to the annotation standards of MS-COCO~\cite{Chen2015cococaption}.

\textbf{Camouflage Disruption}:
To address GPT-4o's difficulty in perceiving camouflaged objects, we propose a camouflage disruption strategy. We extract object contours from COD masks and overlay them on the original images with a distinct color, effectively breaking the camouflage and enhancing object visibility for GPT-4o, as shown in Fig.~\ref{fig:DatasetConstruction} (a). 
It is important to note that the instruction ``ignore the red marks'' was included in the prompt in order to prevent the model from misinterpreting the artificially introduced highlights as an integral part of the scene. 

\textbf{Evolutionary Annotation}:
Given the complexity of describing an image with multiple components like key objects and the environment, it is difficult for a model to generate accurate descriptions for all parts at once. We therefore employ a multi-stage, progressive annotation strategy. As shown in Fig.~\ref{fig:DatasetConstruction} (a), this approach evolves from simple image classification to a description of the camouflaged object, and finally to a comprehensive image description. A key feature of this process is that the category names identified in the initial classification stage are explicitly incorporated into the prompts for subsequent stages, thereby enhancing the quality and accuracy of the final description.

\textbf{Manual Verification and Refinement}:
To ensure high-quality annotations, we incorporate a mandatory manual verification and refinement stage. 
16 trained annotators systematically conducted three rounds of description review and refinement in accordance with standardized guidelines.
Their tasks included (1) correcting factual errors or hallucinations, and (2) improving conciseness by removing redundant content. This process required approximately two to four minutes per image.

CamoIT comprises a total of 10,464 samples, covering approximately 237 categories---including both animals and artificial camouflaged objects---across diverse environments such as marine and terrestrial scenes.
Each image is assigned multi-granularity labels, as shown in Fig.~\ref{fig:DatasetConstruction} (b). 
It is noteworthy that CamoIT's multi-granularity annotations, accompanied by original masks, are suitable for a broad spectrum of tasks beyond ITR. For example, object-level annotations (Level 1 and 2) could benefit community's research in visual grounding and COD~\cite{Zhang2024ACUMEN,Pang2024OVCOS}, though we finally utilize Level 3 annotations for CA-ITR.
The distribution of samples across these categories is visualized in Fig.~\ref{fig:DatasetConstruction} (d), with larger circles closer to the center indicating higher sample frequencies and brighter colors reflecting a greater average number of annotated words per category.  
From Fig.~\ref{fig:DatasetConstruction} (d), it can be observed that the most common object categories in CamoIT correspond to relatively frequent camouflage objects in everyday life, such as lizards and spiders.
The distribution of description lengths, as shown in Fig.~\ref{fig:DatasetConstruction} (c), ranges from 10 to 45 words, with most descriptions clustered around 25 words. 
We divided the samples into training and test sets using a stratified split with an approximate ratio of 7:3 per class, ensuring that both sets maintained a comparable class distribution. For categories with insufficient samples, we perform random allocation while ensuring a balanced distribution of category counts between the training and test sets. The final test set comprises exactly 3,000 samples.

By using CamoIT, we conducted a comprehensive CA-ITR benchmark study; details are in Section~\ref{sec:experiment}.

\section{Camouflage-Expert Collaborative Network}
\label{sec:method}
CA-ITR faces the fundamental challenge of perceiving and representing camouflaged objects. A natural solution is to enhance the target region within a single encoder by employing learnable prompts or COD masks as soft attention mechanisms. However, when highly similar backgrounds interfere with the object during encoding, methods that merely adjust regional weights are fundamentally inadequate to counteract this interference. Although applying masks before encoding can disrupt the camouflage structure and accentuate the target, this strategy risks degrading image fidelity and introducing semantic misalignment, ultimately weakening cross-modal alignment.

\begin{figure*}[!t]
\centering
\includegraphics[width=0.96\linewidth]{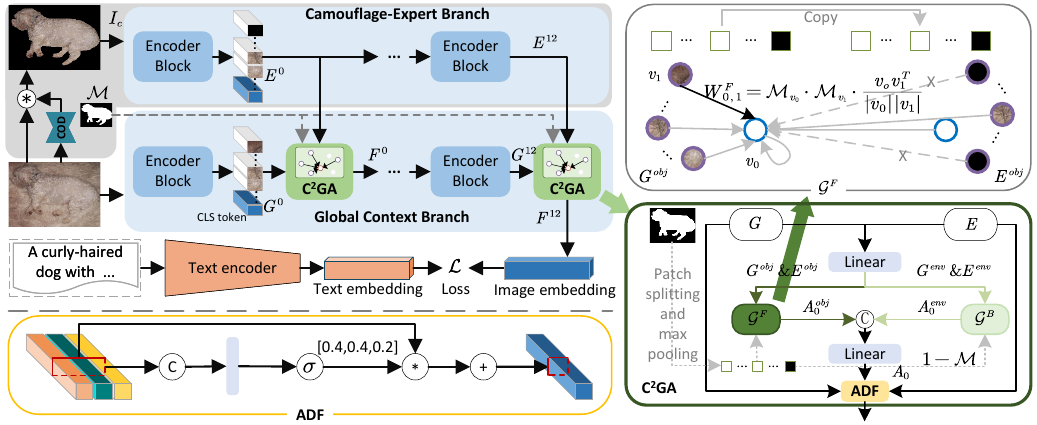}
\vspace{-10pt}
\caption{The overall pipeline of the proposed CECNet and C\textsuperscript{2}GA.
}
\vspace{-15pt}
\label{fig:network}
\end{figure*}

As a solution, we propose a novel camouflage-expert collaborative network (CECNet), which introduces a dual-branch architecture designed to fundamentally prevent feature contamination. One branch processes the original image to preserve global context, while the other independently encodes the isolated camouflaged object to generate a purified feature representation. A novel confidence-conditioned graph attention (C\textsuperscript{2}GA) module is then proposed to intelligently fuse these complementary streams. As illustrated in Fig.~\ref{fig:network}, CECNet is built upon a standard VSE framework for global alignment.
Below, each component will be discussed in detail.

\subsection{Camouflage-Expert Branch} 

Given a camouflaged image $I$, it is processed by a COD method to generate the mask $\mathcal{M}$ of the camouflaged object. This mask is then applied to the original image via element-wise multiplication, yielding a refined image representation $I_c = \mathcal{M} \otimes I$ that primarily preserves the camouflaged object. Subsequently, $I_c$ is fed to   a camouflaged object encoder based on a visual Transformer~\cite{radford2021clip} to extract multi-level features $E^l$, where $l=1, \dots, 12$ indicates the feature level. Each $E^l$ consists of $(N+1)$ tokens: a CLS token $E^l_0$ and $N$ image patch tokens $E^l_i$ ($i = 1, \dots, N$).

\subsection{Global Context Branch}
This branch processes the original image $I$ using a standard vision Transformer to generate multi-level features $G^l$, which effectively capture the holistic scene context. Similarly, $G^l$ comprises $N+1$ tokens: $G^l_i$, where $i = 0, \dots, N$.
To progressively enhance the representation of camouflaged objects, we incorporate the C\textsuperscript{2}GA module (detailed in Sec.~\ref{sec:C2GA}) after each block in the encoder. This allows the global branch to leverage features from the camouflage expert branch, resulting in a more effective global representation. The final feature output of this branch (CLS token) is then used as the ultimate visual representation.

\subsection{Confidence-Conditioned Graph Attention}
\label{sec:C2GA}

Directly fusing features from the two branches using simple strategies like addition or linear transformation is suboptimal (see Sec.~\ref{sec:ablation}). The concern is that these approaches mix all features indiscriminately, risking contamination of the expert branch's purified camouflage representation by dominant background features from the global branch.

To address this issue, we propose a Confidence-Conditioned Graph Attention (C\textsuperscript{2}GA) mechanism, which leverages camouflage confidence scores to separately aggregate foreground and background features, thereby preventing feature contamination and enhancing camouflage representation in a more effective manner. 
C\textsuperscript{2}GA is inspired by multi-head attention mechanisms, which have been shown to effectively capture diverse aspects of input through the construction of distinct subspaces.  
Differently, C\textsuperscript{2}GA explicitly constructs a foreground-relevant graph and a background-relevant graph, guiding the network to focus on each semantic category within its respective subspace.

Specifically, as shown in Fig.~\ref{fig:network}, C\textsuperscript{2}GA takes as input the global branch features $G$ (hereafter, the layer superscript $l$ is omitted for simplicity), the expert branch features $E$, and the mask $\mathcal{M}$. $\mathcal{M}$ is used to compute a camouflage confidence score for each patch feature by applying max pooling to the corresponding region. 
Following multi-head attention, C\textsuperscript{2}GA applies a linear projection to map $G$ into two distinct subspaces, yielding the foreground-oriented representation $G^{obj}$ and the background-oriented representation $G^{env}$. 
Similarly, $E$ is projected through the same linear projection to generate $E^{obj}$ and $E^{env}$.
These representations, along with their corresponding camouflage confidence scores, are utilized to construct a foreground-relevant graph $\mathcal{G}^F$ and a background-relevant graph $\mathcal{G}^B$. The node set of $\mathcal{G}^F$ comprises all tokens from $G^{obj}$ and $E^{obj}$, whereas that of $\mathcal{G}^B$ includes tokens from $G^{env}$ and $E^{env}$.

Taking $\mathcal{G}^F$ as an example, we employ a confidence score to modulate node relationships and suppress edges involving background nodes, thereby constructing a graph focused on foreground-related structures.
In this graph, the edge weight between two nodes $v_i$ and $v_j$ is determined by their similarity and camouflage confidence score:
\begin{align}
W^F_{i,j} = \mathcal{M}_{v_i}\cdot \mathcal{M}_{v_j} \cdot \frac{v_i v_j^T}{|v_i| |v_j|}, \quad v_i, v_j \in \mathcal{V}^F
\end{align}
where $\mathcal{M}_{v_i} \in [0,1]$ and $\mathcal{M}_{v_j} \in [0,1]$ denote the camouflage confidence scores of the 
corresponding nodes. Notably, the camouflage confidence scores for the CLS tokens from $G^{obj}$ and $E^{obj}$ are set to 1, as these tokens contain information about the camouflaged objects.
The camouflage confidence scores of $E^{obj}$ are copied from the node at the corresponding position of $G^{obj}$, as camouflaged objects exhibit consistent representations across both original images.
Once the graph is constructed, foreground-relevant information is aggregated through $\dot{v}_{i} = \sum_{v_j \in \mathcal{V}^F}W^F_{i,j} v_j.$
Note that, due to the use of the global alignment method, we exclusively aggregate information into $G^{obj}_0$ (CLS token) to enhance its feature representation, resulting in $A^{obj}_0$.

Similarly, $\mathcal{G}^B$ is also constructed and performs background-relevant information aggregation, resulting in $A^{env}_0$. In this case, the confidence score of a node $v_j$ from $G^{env}$ is defined as the background confidence score $(1 - \mathcal{M}_{v_j})$, while the confidence score for $G^{env}_0$ is set to 1. The confidence scores of all nodes from $E^{env}$ are assigned 0, as they do not contain any background information.

The enhanced foreground ($A^{obj}_0$) and background ($A^{env}_0$) representations are then concatenated and projected back to the original feature space, achieving enhanced global features $A_0$. To ensure feature stability, we employ adaptive gating fusion \jy{(ADF in Fig.~\ref{fig:network})} to integrate enhanced global features with the original global features:
\begin{align}
    F_{0} = sum(\sigma(f([A_0,E_0, G_0]))\cdot[A_0,E_0, G_0]),
\end{align}
where $f([\cdot])$ utilizes a linear map to adaptively learn the fusion weights for each feature across all channel components, as shown in Fig.~\ref{fig:network}, and $\sigma$ is the sigmoid function. Similarly, by using the global alignment method, we update only the CLS token in the global context branch, leaving the other patch tokens unchanged.

\subsection{Network Optimization}
The visual features extracted from CECNet are aligned with their textual counterparts. Our baseline is built by employing the standard Transformer~\cite{radford2021clip} for text encoding, and the InfoNCE loss~\cite{Van2018InfoNCE} for global cross-modal alignment, i.e. $\mathcal{L} = \mathcal{L}_{T2I} + \mathcal{L}_{I2T}$, 
where $\mathcal{L}_{T2I}$ is text-to-image loss:
\begin{align}
    \mathcal{L}_{T2I} = -\frac{1}{n} \sum_{i=1}^{n} \log \frac{e^{(T_iV_i ^ \top / \tau)}}{\sum_{j=1}^{n} e^{(T_i V_j^\top / \tau)}},
\end{align}
and $\mathcal{L}_{I2T}$ (image-to-text loss) is defined similarly. Here, $V_i$ and $T_i$ denote the visual and textual features of the $i$-th sample in a batch, $n$ is the batch size, and the temperature parameter $\tau$ is set to 0.05.
Notably, as CA-ITR is a very challenging cross-modal alignment task, facing challenges such as bridging the representation gap (where objects are visually similar to the background yet semantically distinct), exploring more elaborate text-side encoding or loss strategies would be a promising direction for future research.

\vspace{-1mm}
\section{Experiments and Results}
\label{sec:experiment}

\begin{table*}[!t]
\centering
\caption{Quantitative results ($R@K$, \%) of models trained on MS-COCO~\cite{lin2014coco} (left) and Flickr30K~\cite{young2014flickr} (right). I2T means image-to-text retrieval, while T2I denotes text-to-image retrieval. ``FG'' and ``BU'' mean whether the model employs local alignment or utilizes the BUTD framework~\cite{Anderson2017butd}. ``N/A'' means that the result is unavailable. The highest scores are marked in \textbf{bold}. ``Pub.'' denotes the publication year.}
\vspace{-8pt}
\label{tab:benchmark}
\tabcolsep 2.2pt
\renewcommand{\arraystretch}{1}
\footnotesize
\begin{tabular*}{0.98\textwidth}{lcccccccccccccccccccc}
\toprule
\multirow{3}{*}[-1.0ex]{Models} & \multirow{3}{*}[-1.0ex]{Pub.} & \multirow{3}{*}[-1.0ex]{FG} & \multirow{3}{*}[-1.0ex]{BU} & \multicolumn{4}{c}{MS-COCO~\cite{lin2014coco}} & \multicolumn{4}{c}{CamoIT} &  & \multicolumn{4}{c}{Flickr30K~\cite{young2014flickr}} & \multicolumn{4}{c}{CamoIT} \\ \cmidrule(lr{0pt}){5-8} \cmidrule(lr{0pt}){9-12} \cmidrule(lr{0pt}){14-17} \cmidrule(lr{0pt}){18-21} 
   & & & & \multicolumn{2}{c}{I2T} & \multicolumn{2}{c}{T2I} & \multicolumn{2}{c}{I2T} & \multicolumn{2}{c}{T2I} & & \multicolumn{2}{c}{I2T} & \multicolumn{2}{c}{T2I} & \multicolumn{2}{c}{I2T} & \multicolumn{2}{c}{T2I}
\\ \cmidrule(lr{0pt}){5-6}  \cmidrule(lr{0pt}){7-8} \cmidrule(lr{0pt}){9-10} \cmidrule(lr{0pt}){11-12} \cmidrule(lr{0pt}){14-15} \cmidrule(lr{0pt}){16-17} \cmidrule(lr{0pt}){18-19} \cmidrule(lr{0pt}){20-21} 
 & & & & $R@1$ & $R@10$ & $R@1$ & $R@10$ & $R@1$ & $R@10$ & $R@1$ & $R@10$ & & $R@1$ & $R@10$ & $R@1$ & $R@10$ & $R@1$ & $R@10$ & $R@1$ & $R@10$\\ 
\midrule
    
CFM~\cite{wei2022cfm}    & 22 &\ding{55} &\ding{51} &59.6&92.9&42.7&83.4&10.7&26.0&10.7&26.7& & 82.7&98.2&62.6&92.0& 5.4&16.5& 6.5 &18.9\\
HREM~\cite{fu2023hrem}   & 23 &\ding{55} &\ding{51} &\textbf{62.5}&\textbf{93.4}&44.0&83.5&11.3&28.1&10.5&27.1& & \textbf{84.8}&97.9&62.4&92.2& 6.7&19.2& 5.7 &17.9\\
CHAN~\cite{pan2023CHAN}  & 23 &\ding{51} &\ding{51} &59.8&93.3&44.9&84.2&10.9&29.7&13.2&30.1& & 80.6&97.8&63.9&92.6& 6.9&20.0& 7.8 &21.7\\
DBL~\cite{diao2024dbl}   & 24 &\ding{51} &\ding{51} &57.5&91.6&41.3&80.9&7.1 &20.0&8.7 &23.5& & 78.2&97.4&58.7&89.5& 4.5&14.2& 3.7 &13.6\\
CUSA~\cite{huang2024cusa}& 24 &\ding{55} &\ding{55} &57.4&90.1&44.3&82.0&\textbf{15.1}&\textbf{37.0}&\textbf{13.5}&\textbf{35.7}& & 81.0&97.9&66.6&94.0 &\textbf{12.6}&\textbf{34.0}&\textbf{10.5} &\textbf{27.6}\\
LAPS~\cite{fu2024laps}   & 24 &\ding{51} &\ding{55} &56.1&91.2&43.9&83.4&11.8&29.8&10.6&27.3& & 75.8&97.2&62.5&92.9& 5.9&17.2& 4.9 &15.7\\
AVSE~\cite{liu2025AVSE}  & 25 &\ding{55} &\ding{55} &N/A &N/A &N/A &N/A &N/A &N/A &N/A &N/A & & 75.9&97.8&62.1&93.0& 5.7&16.0& 4.8 &14.8\\
D2S-VSE~\cite{liu2025D2SVSE} & 25 &\ding{55} &\ding{55} &60.1&92.5&\textbf{46.3}&\textbf{85.2}&13.3&33.1&12.7&30.9& & 83.1&\textbf{98.3}&\textbf{68.5}&\textbf{95.0} & 7.5&20.9& 6.5 &18.6\\
    
\bottomrule
\end{tabular*}
\vspace{-15pt}
\end{table*}

\subsection{Datasets and Protocols}
We conducted a comprehensive benchmark for CA-ITR, evaluating both existing SOTA retrieval methods and our proposed CECNet. 
Our benchmark is conducted on the proposed CamoIT dataset, and two widely recognized benchmark datasets, \ie, the test set of Flickr30K~\cite{young2014flickr} and MS-COCO~\cite{lin2014coco}, are also included for reference.
Following previous methods~\cite{fu2024laps,huang2024cusa,diao2024dbl}, we assess the retrieval performance using recall at $K$ ($R@K$), defined as the proportion of queries for which the correct instance is retrieved among the $K$ nearest neighbors to the query. 
Following MS-COCO and Flickr30K, we use image-level descriptions (Level 3) in CamoIT. 
For a comprehensive evaluation, we use the full test sets: the full 5K set for MS-COCO, the full 3K set for CamoIT, and the full 1K set for Flickr30K.

\vspace{-1mm}
\subsection{Implementation Details}
\vspace{-1mm}
Our CECNet is built upon CLIP (ViT-B/32)~\cite{radford2021clip}, with the text encoder and the global context branch directly adopted from the corresponding components.
The camouflage-expert encoder \emph{shares} parameters with the encoder in the global context branch, thereby ensuring consistent feature representations across both pathways. ZoomNeXt~\cite{Pang2024zoomnext} is selected as the COD expert model.
A two-stage training strategy is employed to optimize the network: C\textsuperscript{2}GA modules are trained initially while maintaining the rest as CLIP weights, followed by end-to-end fine-tuning of the entire CECNet (excluding the COD model).
The overall training process is accelerated using an NVIDIA RTX 4090 GPU.
Following the settings in previous works~\cite{fu2023hrem,radford2021clip}, we use a batch size of 128 and an image size of 224. CECNet is optimized using the Adam optimizer, with the learning rate initially set to 1e-4 in the first stage and subsequently reduced to 1e-5 in the second stage.

The benchmark includes eight SOTA open-source retrieval methods, comprising five global alignment approaches (\ie, CFM~\cite{wei2022cfm}, HREM~\cite{fu2023hrem}, CUSA~\cite{huang2024cusa}, D2S-VSE~\cite{liu2025D2SVSE}, and AVSE~\cite{liu2025AVSE}) and three local alignment approaches (\ie, CHAN~\cite{pan2023CHAN}, DBL~\cite{diao2024dbl}, LAPS~\cite{fu2024laps}). 
Among these methods, CFM, HREM, CHAN, and DBL adopt the SCAN~\cite{Lee2018Stacked} architecture and utilize the frozen bottom-up attention (BUTD)~\cite{Anderson2017butd} for image encoding, whereas the remaining approaches employ trainable visual encoders.
Note that, since only CUSA provides a CLIP-based implementation along with weights, we evaluate its CLIP-based variant to enable a direct comparison with our method. For the other methods (\ie, LAPS, D2S-VSE, AVSE), we evaluate their implementation versions based on the ViT-Base-224~\cite{Dosovitskiy2021ViT} (pre-trained on ImageNet-21K~\cite{Deng2009ImageNet21K}).
To comprehensively evaluate the performance of the aforementioned methods on CA-ITR, we retrain them on CamoIT.

\vspace{-1mm}
\subsection{Benchmark on CA-ITR}
\vspace{-1mm}
\label{sec:benchmark}
Experimental results are summarized in Table~\ref{tab:benchmark} and Table~\ref{tab:benchmark_camoit}. 
As shown in Table~\ref{tab:benchmark}, methods trained on MS-COCO achieve higher performance on CamoIT than those trained on Flickr30K, likely due to the larger scale of MS-COCO (approximately 4 times that of Flickr30K).
Among these methods, CUSA demonstrates superior generalization on CamoIT due to its foundation on CLIP.
However, the performance of all methods, as measured by $R@1$, consistently lies under $15\%$, significantly underperforming relative to their results on MS-COCO and Flickr30K. The results presented in Table~\ref{tab:benchmark_camoit} demonstrate that retraining these models on CamoIT leads to significant performance improvements. 
Nevertheless, even after retraining, all methods underperform compared to their results in conventional retrieval, indicating that \emph{CA-ITR is not simply a domain adaptation task, but one that entails multiple unique challenges.}

It is noteworthy that following retraining, CUSA exhibited only marginal improvement, whereas D2S-VSE demonstrated the most substantial performance gain. The limited performance improvement of CUSA may stem from its use of a frozen image encoder~\cite{An2023Unicom} to correct potential noise in image-text correspondence labels. However, as the encoder is pre-trained on daily scenes, it fails to provide meaningful features for camouflaged scenes, undermining this objective.
In contrast, D2S-VSE emphasizes the key role of information capacity in cross-modal retrieval and introduces a dense text distillation strategy to enhance the information density of sparse text representations. 
Its significant performance gain demonstrates the crucial role of information capacity in CA-ITR and, more fundamentally, \emph{reveals the high complexity of image contents (as can be observed in  Fig.~\ref{fig:qualitative_other}), emphasizing the need for a more granular understanding of CA-ITR} through richer textual descriptions to enable accurate modeling.

\begin{table}[t]
\centering
\caption{Quantitative results ($R@K$, \%) on CA-ITR. All models are \emph{retrained} on CamoIT. The highest scores are marked in \textbf{bold}. 
}
\vspace{-8pt}
\label{tab:benchmark_camoit}
\tabcolsep 4pt
\renewcommand{\arraystretch}{0.9}
\footnotesize
\begin{tabular*}{0.46\textwidth}{lcccccc}
\toprule
    \multirow{2}{*}[-1.0ex]{Models} &  \multicolumn{3}{c}{I2T} & \multicolumn{3}{c}{T2I} 
\\ \cmidrule(lr{0pt}){2-4}  \cmidrule(lr{0pt}){5-7} 
    & $R@1$ & $R@5$ & $R@10$ & $R@1$ & $R@5$ & $R@10$\\ 
    \hline
    CFM~\cite{wei2022cfm} 
        & 30.8 &59.9 &70.3 &28.9 &57.2 &68.3 \\
    HREM~\cite{fu2023hrem} 
        & 34.3 &62.7 &74.0 &31.5 &59.9 &71.4 \\
    CUSA~\cite{huang2024cusa} 
        & 23.9 &53.5 &66.7 &23.5 &51.3 &64.3 \\
    LAPS~\cite{fu2024laps} 
        & 27.8 &62.0 &73.9 &28.2 &58.0 &70.7 \\

    D2S-VSE~\cite{liu2025D2SVSE} 
        & 37.1 &68.4 &79.5 &35.5 &67.5 &78.4 \\
    AVSE~\cite{liu2025AVSE} 
        & 28.1 &59.7 &72.2 &26.1 &56.3 &69.7 \\
    CLIP~\cite{radford2021clip} 
        & 41.3 &69.2 &79.0 &41.1 &67.7 &78.4 \\
        \rowcolor{gray!20}
    D2S-VSE + CEC
        & 39.0 &69.4 &81.0 &37.0 &68.3 &79.7 \\
        \rowcolor{gray!20}
    AVSE + CEC
        & 29.9 &60.7 &73.6 &28.5 &59.2 &71.0 \\
        \rowcolor{gray!20}
    CECNet (Ours)
        & \textbf{45.8} &\textbf{74.5} &\textbf{83.5} &\textbf{44.6} &\textbf{73.9} &\textbf{83.1} \\ 
    
\bottomrule
\end{tabular*}
\vspace{-15pt}
\end{table}

\begin{figure*}[t]
\centering
\includegraphics[width=0.98\linewidth]{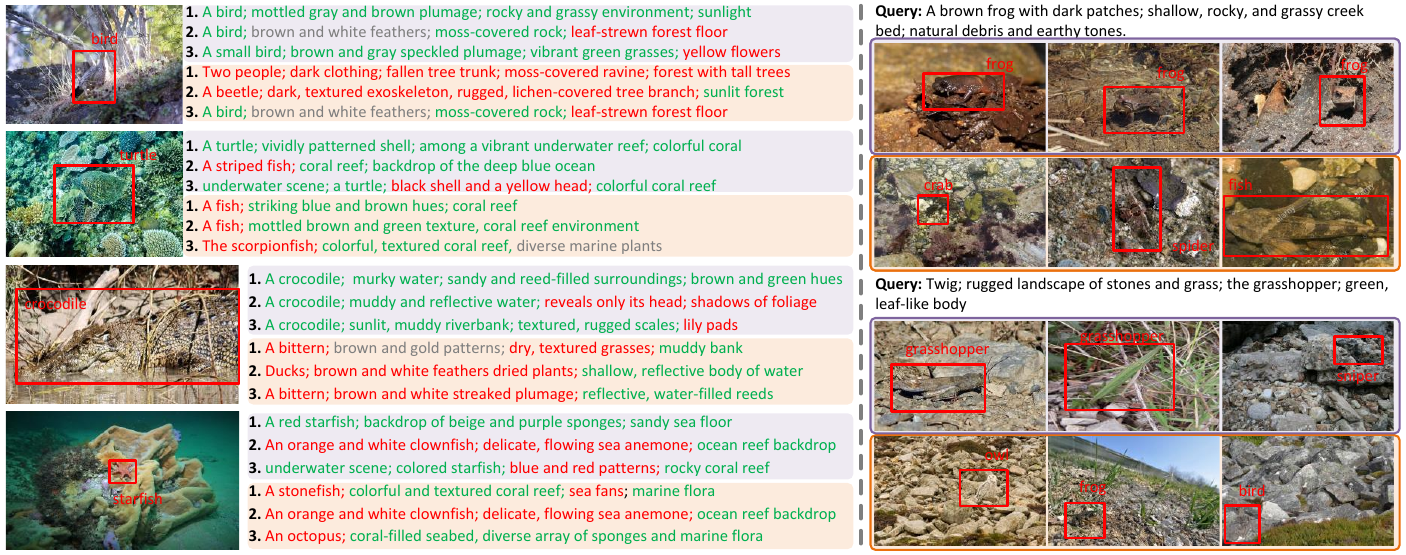}
\vspace{-5pt}
\caption{Qualitative results of CECNet (top, light purple) and CLIP (bottom, light orange) on sentence retrieval (left) and image retrieval (right). For each query, we present the top three relevant cross-modal instances. To enhance readability, we retain only the essential components of the sentence. The accurate, inaccurate, and ambiguous portions of the sentences in the search results are highlighted in green, red, and gray, respectively. Camouflaged objects are marked with red bounding boxes.}
\vspace{-15pt}
\label{fig:qualitative}
\end{figure*}

More interestingly, methods based on the frozen BUTD framework (CFM, HREM) outperform several trainable ViT-based approaches (CUSA, LAPS, AVSE) on CamoIT. This performance advantage can be attributed to BUTD’s object-level representation paradigm, which decomposes an image into explicit object regions and encodes them accordingly. During object proposal generation, BUTD may capture camouflaged objects---even if only partially---particularly when the camouflage is not highly effective, thereby alleviating some of the perceptual challenges associated with such objects. In contrast, ViT relies on holistic features that are entangled with background context, making it more difficult to isolate object-specific signals from uniform or cluttered surroundings. This finding not only demonstrates \emph{the distinct challenges associated with camouflaged object perception in CA-ITR}, but also substantiates the design rationale behind CECNet’s dual-branch architecture and its explicit modeling of camouflaged objects. 
However, BUTD struggles to use background information~\cite{Luo2021BUTDFailure}, which actually highlights the rationale for CECNet's integration of the global context branch.

Overall, the benchmark results demonstrate that CA-ITR is not merely a domain transfer task but one that presents several distinct challenges, including the difficulty of perceiving camouflaged objects, complexity of image contents, and necessity for fine-grained understanding.

\vspace{-1mm}
\subsection{Comparisons with State-of-the-Arts}
\vspace{-1mm}
The quantitative results of CECNet and fine-tuned CLIP are presented in Table~\ref{tab:benchmark_camoit}.
On the benchmark, our CECNet achieves an average R@1 improvement of $8.9\%$ over the strongest general retrieval model, D2S-VSE. This superiority is consistent across different models: CECNet advances the fine-tuned CLIP by an average of $4\%$ and notably surpasses another CLIP-based model, CUSA, by a large margin of $21.5\%$.
\jy{Also, we further apply the camouflage-expert collaborative scheme to D2S-VSE and AVSE (``D2S-VSE + CEC'' and ``AVSE + CEC'' in Table~\ref{tab:benchmark_camoit}), resulting in performance enhancements.}
The consistent and significant gains demonstrate that CECNet effectively tackles the specific challenges of CA-ITR.
As a pioneering exploration in CA-ITR, the primary contribution of CECNet lies in establishing a foundational baseline and validating the feasibility of the task. The achieved performance, which mirrors the early development phase of traditional ITR~\cite{HuangWW2017SMLSTM,EisenschtatW2017ttwoWayNets,VendrovKFU2015OrderEmbeddings}, exemplified by $38.1\%$ on $R@1$ on Flickr30K, should be viewed as a promising starting point that enables and motivates future research in this challenging domain.

We further present qualitative results of CECNet and CLIP on both sentence retrieval and image retrieval tasks, as shown in Fig.~\ref{fig:qualitative}. The results demonstrate that CECNet achieves retrieval outcomes with higher semantic relevance to the query, especially in cross-modal scenarios where the primary objectives closely align with the query. In contrast, although CLIP retrieves results that are strongly tied to the background semantics of the query, the primary objectives in these results frequently mismatch the queried content.
The limitations of CLIP highlight the unique challenges in camouflaged object perception within CA-ITR, whereas CECNet achieves superior performance by enhancing the perception and understanding of camouflaged objects.

\vspace{-1mm}
\subsection{Ablation Study}
\vspace{-1mm}
\label{sec:ablation}

\textbf{Rationality Behind the Dual-Branch Architecture.}
We conducted comprehensive ablation studies to validate the necessity of dual-branch encoder. As summarized in Table~\ref{tab:ablation_on_fusion}, using the standard CLIP model as the baseline (``Baseline''), We evaluated CECNet against several variants specifically designed to enhance attention toward camouflaged object regions within a single encoder. Specifically, ``A1'' applies a mask to adaptively modulate the input image via convolutional operations before encoding; ``A2'' integrates the mask with image features through convolutional fusion; ``A3'' introduces a trainable prompt to learn feature representations of camouflaged objects, computing the similarity between the prompt and image features to generate a class activation mapping (CAM)-like attention map, which is supervised by the ground-truth segmentation mask during training.
Results show that, apart from ``A1'', which causes performance degradation, all other schemes achieve performance improvements, with CECNet demonstrating the most substantial enhancement. These results are consistent with the analyses presented in Section~\ref{sec:method}, thereby confirming the effectiveness and necessity of the dual-branch encoder, and further supporting the rationality of its architectural design as discussed in Section~\ref{sec:benchmark}.

\begin{table}[t]
\centering
\caption{Ablation study of $R@K$ (\%) on dual-branch visual encoder and C\textsuperscript{2}GA. The highest scores are marked in \textbf{bold}.}
\vspace{-8pt}
\label{tab:ablation_on_fusion}
\tabcolsep 6.2pt
\renewcommand{\arraystretch}{0.9}
\footnotesize
\begin{tabular*}{0.47\textwidth}{lcccccc}
\toprule
\multirow{3}{*}[-1.0ex]{Settings} & \multicolumn{6}{c}{CamoIT}
\\ \cmidrule(lr{0pt}){2-7}
    &  \multicolumn{3}{c}{I2T} & \multicolumn{3}{c}{T2I} 
\\ \cmidrule(lr{0pt}){2-4}  \cmidrule(lr{0pt}){5-7}
    &  R@1 & R@5 & R@10  & R@1 & R@5  & R@10 \\ 
    \hline
Baseline     & 41.3 & 69.2 & 79.0 & 41.1 & 67.7 & 78.4 \\
A1          & 28.5 & 58.3 & 70.9 & 30.0 & 58.0 & 70.4 \\
A2          & 42.1 & 69.9 & 79.2 & 41.2 & 69.5 & 79.0 \\ 

A3          & 41.8 & 69.2 & 79.3 & 42.2 & 68.9 & 79.5 \\ 

B1           & 42.5 & 71.7 & 80.7 & 42.9 & 71.1 & 80.3  \\ 
B2           & 42.4 & 70.4 & 80.7 & 41.7 & 70.2 & 79.9  \\
B3   & 42.9 & 70.7 & 80.6 & 42.3 & 68.9 & 78.8  \\
\rowcolor{gray!20}
CECNet              & \textbf{45.8} &\textbf{74.5} &\textbf{83.5} &\textbf{44.6} &\textbf{73.9} &\textbf{83.1}\\ 
\bottomrule
\end{tabular*}
\vspace{-15pt}
\end{table}

\textbf{Effects of C\textsuperscript{2}GA.}
We compared C\textsuperscript{2}GA against several fusion strategies, including a simple addition, linear fusion (via channel concatenation followed by a linear layer), and a vanilla graph attention mechanism, denoted as ``B1'', ``B2'', and ``B3'', respectively. As summarized in Table~\ref{tab:ablation_on_fusion}, both alternatives yield performance gains; however, their improvements are limited relative to C\textsuperscript{2}GA.
The alternative solutions improve performance by leveraging camouflage object information. However, they indiscriminately integrate all features, potentially allowing dominant background features to corrupt the expert branch’s refined camouflage representation.
In contrast, C\textsuperscript{2}GA selectively integrates camouflage object information into the global representation, preventing feature contamination and enhancing the camouflaged object’s representation.

\textbf{Influence of Different COD Models.} 
Our method employs a COD model as a dedicated expert to provide knowledge for camouflaged scene understanding. To evaluate how this expert guidance influences performance, we integrate four COD backbones: SINet~\cite{Fan2020COD10K}, SINet-v2~\cite{fan2022concealed}, ZoomNet~\cite{pang2022zoom}, and ZoomNeXt~\cite{Pang2024zoomnext}. 
We establish performance bounds using all-white images (``White'') to simulate scenarios where object perception is unavailable. 
All COD models are retrained on the CamoIT training set to prevent potential information leakage from prior training data appearing in the test set. We follow each method by initializing training directly from pre-trained weights (e.g., ImageNet~\cite{Deng2009ImageNet21K}).
As shown in Table~\ref{tab:ablation_on_COD}, CECNet effectively translates improvements in expert capability into enhanced overall performance, which validate CECNet's ability to leverage expert knowledge. 
\emph{By the way, our validation is the first to verify that incorporating COD models into the image-text retrieval paradigm can truly boost the retrieval accuracy.} 

\begin{table}[!t]
\centering
\caption{Retrieval performance of CECNet and segmentation performance of corresponding COD models. ``White'' denotes the replacement of COD detection results with a white image. 
Segmentation performance is measured by $S_\alpha$ and $MAE$ (see~\cite{Fan2017StructureMeasureAN}).
}
\vspace{-8pt}
\label{tab:ablation_on_COD}
\tabcolsep 4.7pt
\renewcommand{\arraystretch}{0.95}
\footnotesize
\begin{tabular*}{0.47\textwidth}{lcccccc}
\toprule
\multirow{3}{*}[-1.0ex]{COD Models} & \multicolumn{4}{c}{Retrieval} & \multicolumn{2}{c}{Segmentation}
\\ \cmidrule(lr{0pt}){2-5}  \cmidrule(lr{0pt}){6-7} 
    &  \multicolumn{2}{c}{I2T} & \multicolumn{2}{c}{T2I} & \multirow{2}{*}[-1.0ex]{$S_\alpha$}&\multirow{2}{*}[-1.0ex]{$MAE$}
\\ \cmidrule(lr{0pt}){2-3}  \cmidrule(lr{0pt}){4-5}
    &  R@1 & R@5  & R@1  & R@5  \\ \hline
White                                 & 41.6 & 69.6 & 41.3 &69.4 &  0.443    &  0.120  \\
SINet~\cite{Fan2020COD10K}            & 42.3 & 70.4 & 42.1 &69.9 &  0.808    &  0.049  \\
SINet-v2~\cite{fan2022concealed}      & 43.3 & 72.7 & 43.1 &71.7 &  0.843    &  0.037  \\
ZoomNet~\cite{pang2022zoom}           & 44.0 & 71.6 & 42.9 &70.5 &  0.851    &  0.033  \\
\rowcolor{gray!20}
ZoomNeXt~\cite{Pang2024zoomnext}      & \textbf{45.8} & \textbf{74.5} & \textbf{44.6} &\textbf{73.9} &  \textbf{0.906}    &  \textbf{0.021}  \\
\bottomrule
\end{tabular*}
\vspace{-15pt}
\end{table}
\vspace{-1mm}
\section{Conclusion}
\vspace{-1mm}
\label{sec:conclusion}

This paper is the first to study cross-modal retrieval in camouflaged scenarios, formally defining the task of camouflage-aware image-text retrieval (CA-ITR). We construct a dedicated camouflage image-text retrieval dataset (CamoIT) as support, and the benchmark conducted on CamoIT reveals the underlying challenges of CA-ITR. To address CA-ITR, we propose a camouflage-expert collaborative network (CECNet), which incorporates a COD expert and a novel confidence-conditioned graph attention (C²GA) mechanism into a dual-branch visual encoder to enhance camouflaged object representations. Extensive experiments show that CECNet achieves encouraging performance over seven representative retrieval models.

\vspace{-2pt}
\small{\vspace{.1in}\noindent\textbf{Acknowledgments.}\quad
\jy{This work was supported by the NSFC, under No. 62176169, and the Sichuan Science and Technology Program (2025ZNSFSC0469).}}

{
    \small
    \bibliographystyle{ieeenat_fullname}
    \bibliography{main}

@String(PAMI = {IEEE Trans. Pattern Anal. Mach. Intell.})

@String(CVPR= {IEEE Conf. Comput. Vis. Pattern Recog.})

@String(ICCV= {Int. Conf. Comput. Vis.})

@String(ECCV= {Eur. Conf. Comput. Vis.})

@String(NIPS= {Adv. Neural Inform. Process. Syst.})

@String(BMVC= {Brit. Mach. Vis. Conf.})

@String(TIP  = {IEEE Trans. Image Process.})

@String(ACMMM= {ACM Int. Conf. Multimedia})

@String(ICLR = {Int. Conf. Learn. Represent.})

@String(IJCAI = {IJCAI})

@String(AAAI = {AAAI})

@String(PAMI  = {IEEE TPAMI})

@String(CVPR  = {CVPR})

@String(ICCV  = {ICCV})

@String(ECCV  = {ECCV})

@String(NIPS  = {NeurIPS})

@String(BMVC  =	{BMVC})

@String(TIP   = {IEEE TIP})

@String(TCSVT = {IEEE TCSVT})

@String(ACMMM = {ACM MM})

@String(ICLR  = {ICLR})

@inproceedings{radford2021clip,
  author       = {Alec Radford and
                  Jong Wook Kim and
                  Chris Hallacy and
                  Aditya Ramesh and
                  Gabriel Goh and
                  Sandhini Agarwal and
                  Girish Sastry and
                  Amanda Askell and
                  Pamela Mishkin and
                  Jack Clark and
                  Gretchen Krueger and
                  Ilya Sutskever},
  title        = {Learning Transferable Visual Models From Natural Language Supervision},
  booktitle    = {ICML},
  volume       = {139},
  pages        = {8748--8763},
  year         = {2021}
}

@article{jiang2024effectiveness,
  author       = {Yao Jiang and
                  Xinyu Yan and
                  Ge{-}Peng Ji and
                  Keren Fu and
                  Meijun Sun and
                  Huan Xiong and
                  Deng{-}Ping Fan and
                  Fahad Shahbaz Khan},
  title        = {Effectiveness assessment of recent large vision-language models},
  journal      = {Vis. Intell.},
  volume       = {2},
  number       = {1},
  pages        = {17},
  year         = {2024}
}

@article{le2019anabranch,
  author       = {Trung{-}Nghia Le and
                  Tam V. Nguyen and
                  Zhongliang Nie and
                  Minh{-}Triet Tran and
                  Akihiro Sugimoto},
  title        = {Anabranch network for camouflaged object segmentation},
  journal      = {Comput. Vis. Image Underst.},
  volume       = {184},
  pages        = {45--56},
  year         = {2019}
}

@article{fan2022concealed,
  author       = {Deng{-}Ping Fan and
                  Ge{-}Peng Ji and
                  Ming{-}Ming Cheng and
                  Ling Shao},
  title        = {Concealed Object Detection},
  journal      = PAMI,
  volume       = {44},
  number       = {10},
  pages        = {6024--6042},
  year         = {2022}
}

@article{bi2022rethinkingcod,
  author       = {Hongbo Bi and
                  Cong Zhang and
                  Kang Wang and
                  Jinghui Tong and
                  Feng Zheng},
  title        = {Rethinking Camouflaged Object Detection: Models and Datasets},
  journal      = TCSVT,
  volume       = {32},
  number       = {9},
  pages        = {5708--5724},
  year         = {2022}
}

@article{fan2023advancescod,
  author       = {Deng{-}Ping Fan and
                  Ge{-}Peng Ji and
                  Peng Xu and
                  Ming{-}Ming Cheng and
                  Christos Sakaridis and
                  Luc Van Gool},
  title        = {Advances in deep concealed scene understanding},
  journal      = {Vis. Intell.},
  volume       = {1},
  number       = {1},
  year         = {2023}
}

@inproceedings{fan2020pranet,
  author       = {Deng{-}Ping Fan and
                  Ge{-}Peng Ji and
                  Tao Zhou and
                  Geng Chen and
                  Huazhu Fu and
                  Jianbing Shen and
                  Ling Shao},
  title        = {PraNet: Parallel Reverse Attention Network for Polyp Segmentation},
  booktitle    = {MICCAI},
  volume       = {12266},
  pages        = {263--273},
  year         = {2020}
}

@inproceedings{wei2022cfm,
  author       = {Hao Wei and
                  Shuhui Wang and
                  Xinzhe Han and
                  Zhe Xue and
                  Bin Ma and
                  Xiaoming Wei and
                  Xiaolin Wei},
  title        = {Synthesizing Counterfactual Samples for Effective Image-Text Matching},
  booktitle    = ACMMM,
  pages        = {4355--4364},
  year         = {2022}
}

@inproceedings{fu2023hrem,
  author       = {Zheren Fu and
                  Zhendong Mao and
                  Yan Song and
                  Yongdong Zhang},
  title        = {Learning Semantic Relationship among Instances for Image-Text Matching},
  booktitle    = CVPR,
  pages        = {15159--15168},
  year         = {2023}
}

@article{diao2024dbl,
  author       = {Haiwen Diao and
                  Ying Zhang and
                  Shang Gao and
                  Xiang Ruan and
                  Huchuan Lu},
  title        = {Deep Boosting Learning: {A} Brand-New Cooperative Approach for Image-Text
                  Matching},
  journal      = TIP,
  volume       = {33},
  pages        = {3341--3352},
  year         = {2024}
}

@inproceedings{huang2024cusa,
  author       = {Hailang Huang and
                  Zhijie Nie and
                  Ziqiao Wang and
                  Ziyu Shang},
  title        = {Cross-Modal and Uni-Modal Soft-Label Alignment for Image-Text Retrieval},
  booktitle    = AAAI,
  pages        = {18298--18306},
  year         = {2024}
}

@inproceedings{fu2024laps,
  author       = {Zheren Fu and
                  Lei Zhang and
                  Hou Xia and
                  Zhendong Mao},
  title        = {Linguistic-Aware Patch Slimming Framework for Fine-Grained Cross-Modal
                  Alignment},
  booktitle    = CVPR,
  pages        = {26297--26306},
  year         = {2024}
}

@inproceedings{lin2014coco,
  author       = {Tsung{-}Yi Lin and
                  Michael Maire and
                  Serge J. Belongie and
                  James Hays and
                  Pietro Perona and
                  Deva Ramanan and
                  Piotr Doll{\'{a}}r and
                  C. Lawrence Zitnick},
  title        = {Microsoft {COCO:} Common Objects in Context},
  booktitle    = ECCV,
  volume       = {8693},
  pages        = {740--755},
  year         = {2014}
}

@article{young2014flickr,
  author       = {Peter Young and
                  Alice Lai and
                  Micah Hodosh and
                  Julia Hockenmaier},
  title        = {From image descriptions to visual denotations: New similarity metrics
                  for semantic inference over event descriptions},
  journal      = {Trans. Assoc. Comput. Linguistics},
  volume       = {2},
  pages        = {67--78},
  year         = {2014}
}

@article{yan2021mirrornet,
  author       = {Jinnan Yan and
                  Trung{-}Nghia Le and
                  Khanh{-}Duy Nguyen and
                  Minh{-}Triet Tran and
                  Thanh{-}Toan Do and
                  Tam V. Nguyen},
  title        = {MirrorNet: Bio-Inspired Camouflaged Object Segmentation},
  journal      = {{IEEE} Access},
  volume       = {9},
  pages        = {43290--43300},
  year         = {2021}
}

@inproceedings{wu2023sourcefree,
  author       = {Zongwei Wu and
                  Danda Pani Paudel and
                  Deng{-}Ping Fan and
                  Jingjing Wang and
                  Shuo Wang and
                  C{\'{e}}dric Demonceaux and
                  Radu Timofte and
                  Luc Van Gool},
  title        = {Source-free Depth for Object Pop-out},
  booktitle    = ICCV,
  pages        = {1032--1042},
  year         = {2023}
}

@inproceedings{zhong2022codfrequency,
  author       = {Yijie Zhong and
                  Bo Li and
                  Lv Tang and
                  Senyun Kuang and
                  Shuang Wu and
                  Shouhong Ding},
  title        = {Detecting Camouflaged Object in Frequency Domain},
  booktitle    = CVPR,
  pages        = {4494--4503},
  year         = {2022}
}

@inproceedings{pang2022zoom,
  author       = {Youwei Pang and
                  Xiaoqi Zhao and
                  Tian{-}Zhu Xiang and
                  Lihe Zhang and
                  Huchuan Lu},
  title        = {Zoom In and Out: {A} Mixed-scale Triplet Network for Camouflaged Object
                  Detection},
  booktitle    = CVPR,
  pages        = {2150--2160},
  year         = {2022}
}

@inproceedings{zheng2023mffn,
  author       = {Dehua Zheng and
                  Xiaochen Zheng and
                  Laurence T. Yang and
                  Yuan Gao and
                  Chenlu Zhu and
                  Yiheng Ruan},
  title        = {{MFFN:} Multi-view Feature Fusion Network for Camouflaged Object Detection},
  booktitle    = {WACV},
  pages        = {6221--6231},
  year         = {2023}
}

@inproceedings{Kajiura2021Improving,
  author       = {Nobukatsu Kajiura and
                  Hong Liu and
                  Shin'ichi Satoh},
  title        = {Improving Camouflaged Object Detection with the Uncertainty of Pseudo-edge
                  Labels},
  booktitle    = {ACM Multimedia Asia},
  pages        = {7:1--7:7},
  year         = {2021}
}

@inproceedings{sun2021contextaware,
  author       = {Yujia Sun and
                  Geng Chen and
                  Tao Zhou and
                  Yi Zhang and
                  Nian Liu},
  title        = {Context-aware Cross-level Fusion Network for Camouflaged Object Detection},
  booktitle    = {IJCAI},
  pages        = {1025--1031},
  year         = {2021}
}

@inproceedings{jia2022Segment,
  author       = {Qi Jia and
                  Shuilian Yao and
                  Yu Liu and
                  Xin Fan and
                  Risheng Liu and
                  Zhongxuan Luo},
  title        = {Segment, Magnify and Reiterate: Detecting Camouflaged Objects the
                  Hard Way},
  booktitle    = CVPR,
  pages        = {4703--4712},
  year         = {2022}
}

@inproceedings{mei2021distractionmining,
  author       = {Haiyang Mei and
                  Ge-Peng Ji and
                  Ziqi Wei and
                  Xin Yang and
                  Xiaopeng Wei and
                  Deng-Ping Fan},
  title        = {Camouflaged Object Segmentation With Distraction Mining},
  booktitle    = CVPR,
  pages        = {8772--8781},
  year         = {2021}
}

@article{mei2023distraction,
  title={Distraction-aware camouflaged object segmentation},
  author={Mei, Haiyang and Yang, Xin and Zhou, Yunduo and Ji, Ge-Peng and Wei, Xiaopeng and Fan, Deng-Ping},
  journal={SCIENTIA SINICA Informationis},
  volume={3},
  number={7},
  year={2023}
}

@inproceedings{huang2023shrinkage,
  author       = {Zhou Huang and
                  Hang Dai and
                  Tian{-}Zhu Xiang and
                  Shuo Wang and
                  Huai{-}Xin Chen and
                  Jie Qin and
                  Huan Xiong},
  title        = {Feature Shrinkage Pyramid for Camouflaged Object Detection with Transformers},
  booktitle    = CVPR,
  pages        = {5557--5566},
  year         = {2023}
}

@inproceedings{li2021uncertaintyaware,
  author       = {Aixuan Li and
                  Jing Zhang and
                  Yunqiu Lv and
                  Bowen Liu and
                  Tong Zhang and
                  Yuchao Dai},
  title        = {Uncertainty-Aware Joint Salient Object and Camouflaged Object Detection},
  booktitle    = CVPR,
  year         = {2021}
}

@inproceedings{frome2013DeViSE,
  author       = {Andrea Frome and
                  Gregory S. Corrado and
                  Jonathon Shlens and
                  Samy Bengio and
                  Jeffrey Dean and
                  Marc'Aurelio Ranzato and
                  Tom{\'{a}}s Mikolov},
  title        = {DeViSE: {A} Deep Visual-Semantic Embedding Model},
  booktitle    = NIPS,
  pages        = {2121--2129},
  year         = {2013}
}

@inproceedings{Faghri2018vsepp,
  author       = {Fartash Faghri and
                  David J. Fleet and
                  Jamie Ryan Kiros and
                  Sanja Fidler},
  title        = {{VSE++:} Improving Visual-Semantic Embeddings with Hard Negatives},
  booktitle    = {BMVC},
  pages        = {12},
  year         = {2018}
}

@inproceedings{Sarafianos2019Adversarial,
  author       = {Nikolaos Sarafianos and
                  Xiang Xu and
                  Ioannis A. Kakadiaris},
  title        = {Adversarial Representation Learning for Text-to-Image Matching},
  booktitle    = {ICCV},
  pages        = {5813--5823},
  year         = {2019}
}

@article{li2023Reasoning,
  author       = {Kunpeng Li and
                  Yulun Zhang and
                  Kai Li and
                  Yuanyuan Li and
                  Yun Fu},
  title        = {Image-Text Embedding Learning via Visual and Textual Semantic Reasoning},
  journal      = PAMI,
  volume       = {45},
  number       = {1},
  pages        = {641--656},
  year         = {2023}
}

@article{zhang2024user,
  author       = {Yan Zhang and
                  Zhong Ji and
                  Di Wang and
                  Yanwei Pang and
                  Xuelong Li},
  title        = {{USER:} Unified Semantic Enhancement With Momentum Contrast for Image-Text
                  Retrieval},
  journal      = TIP,
  volume       = {33},
  pages        = {595--609},
  year         = {2024}
}

@inproceedings{Lee2018Stacked,
  author       = {Kuang{-}Huei Lee and
                  Xi Chen and
                  Gang Hua and
                  Houdong Hu and
                  Xiaodong He},
  title        = {Stacked Cross Attention for Image-Text Matching},
  booktitle    = ECCV,
  volume       = {11208},
  pages        = {212--228},
  year         = {2018}
}

@inproceedings{Wang2019camp,
  author       = {Zihao Wang and
                  Xihui Liu and
                  Hongsheng Li and
                  Lu Sheng and
                  Junjie Yan and
                  Xiaogang Wang and
                  Jing Shao},
  title        = {{CAMP:} Cross-Modal Adaptive Message Passing for Text-Image Retrieval},
  booktitle    = ICCV,
  pages        = {5763--5772},
  year         = {2019}
}

@inproceedings{Liu2020Graph,
  author       = {Chunxiao Liu and
                  Zhendong Mao and
                  Tianzhu Zhang and
                  Hongtao Xie and
                  Bin Wang and
                  Yongdong Zhang},
  title        = {Graph Structured Network for Image-Text Matching},
  booktitle    = CVPR,
  pages        = {10918--10927},
  year         = {2020}
}

@article{Li2024MultimodalAlignmentSurvey,
  author       = {Songtao Li and
                  Hao Tang},
  title        = {Multimodal Alignment and Fusion: {A} Survey},
  journal      = {CoRR},
  volume       = {abs/2411.17040},
  year         = {2024}
}

@article{Tadas2019SurveyandTaxonomy,
  author       = {Tadas Baltrusaitis and
                  Chaitanya Ahuja and
                  Louis{-}Philippe Morency},
  title        = {Multimodal Machine Learning: {A} Survey and Taxonomy},
  journal      = PAMI,
  volume       = {41},
  number       = {2},
  pages        = {423--443},
  year         = {2019}
}

@inproceedings{Li2021ALBEF,
  author       = {Junnan Li and
                  Ramprasaath R. Selvaraju and
                  Akhilesh Gotmare and
                  Shafiq R. Joty and
                  Caiming Xiong and
                  Steven Chu{-}Hong Hoi},
  title        = {Align before Fuse: Vision and Language Representation Learning with
                  Momentum Distillation},
  booktitle    = NIPS,
  pages        = {9694--9705},
  year         = {2021},
}

@article{hurst2024gpt4o,
  title={Gpt-4o system card},
  author={Hurst, Aaron and Lerer, Adam and Goucher, Adam P and Perelman, Adam and Ramesh, Aditya and Clark, Aidan and Ostrow, AJ and Welihinda, Akila and Hayes, Alan and Radford, Alec and others},
  journal={arXiv preprint arXiv:2410.21276},
  year={2024}
}

@article{skurowski2018CHAMELEON,
  title={Animal camouflage analysis: Chameleon database},
  author={Skurowski, Przemys{\l}aw and Abdulameer, Hassan and B{\l}aszczyk, Jakub and Depta, Tomasz and Kornacki, Adam and Kozie{\l}, Przemys{\l}aw},
  journal={Unpublished manuscript},
  volume={2},
  number={6},
  pages={7},
  year={2018}
}

@inproceedings{Fan2020COD10K,
  author       = {Deng{-}Ping Fan and
                  Ge{-}Peng Ji and
                  Guolei Sun and
                  Ming{-}Ming Cheng and
                  Jianbing Shen and
                  Ling Shao},
  title        = {Camouflaged Object Detection},
  booktitle    = CVPR,
  pages        = {2774--2784},
  year         = {2020}
}

@inproceedings{Lv2021NC4K,
  author       = {Yunqiu Lv and
                  Jing Zhang and
                  Yuchao Dai and
                  Aixuan Li and
                  Bowen Liu and
                  Nick Barnes and
                  Deng{-}Ping Fan},
  title        = {Simultaneously Localize, Segment and Rank the Camouflaged Objects},
  booktitle    = CVPR,
  pages        = {11591--11601},
  year         = {2021}
}

@inproceedings{Anderson2017butd,
  author       = {Peter Anderson and
                  Xiaodong He and
                  Chris Buehler and
                  Damien Teney and
                  Mark Johnson and
                  Stephen Gould and
                  Lei Zhang},
  title        = {Bottom-Up and Top-Down Attention for Image Captioning and Visual Question Answering},
  booktitle    = CVPR,
  year={2018},
  pages={6077-6086},
}

@inproceedings{An2023Unicom,
  author       = {Xiang An and
                  Jiankang Deng and
                  Kaicheng Yang and
                  Jaiwei Li and
                  Ziyong Feng and
                  Jia Guo and
                  Jing Yang and
                  Tongliang Liu},
  title        = {Unicom: Universal and Compact Representation Learning for Image Retrieval},
  booktitle    = ICLR,
  year         = {2023}
}

@article{Pang2024zoomnext,
  author       = {Youwei Pang and
                  Xiaoqi Zhao and
                  Tian{-}Zhu Xiang and
                  Lihe Zhang and
                  Huchuan Lu},
  title        = {ZoomNeXt: {A} Unified Collaborative Pyramid Network for Camouflaged
                  Object Detection},
  journal      = PAMI,
  volume       = {46},
  number       = {12},
  pages        = {9205--9220},
  year         = {2024}
}

@article{Li2023ZSCOD,
  author       = {Haoran Li and
                  Chun{-}Mei Feng and
                  Yong Xu and
                  Tao Zhou and
                  Lina Yao and
                  Xiaojun Chang},
  title        = {Zero-Shot Camouflaged Object Detection},
  journal      = TIP,
  volume       = {32},
  pages        = {5126--5137},
  year         = {2023}
}

@inproceedings{Tang2024MMCPF,
  author       = {Lv Tang and
                  Peng{-}Tao Jiang and
                  Zhihao Shen and
                  Hao Zhang and
                  Jin{-}Wei Chen and
                  Bo Li},
  title        = {Chain of Visual Perception: Harnessing Multimodal Large Language Models
                  for Zero-shot Camouflaged Object Detection},
  booktitle    = ACMMM,
  pages        = {8805--8814},
  year         = {2024}
}

@inproceedings{Hu2024GenSAM,
  author       = {Jian Hu and
                  Jiayi Lin and
                  Shaogang Gong and
                  Weitong Cai},
  title        = {Relax Image-Specific Prompt Requirement in {SAM:} {A} Single Generic
                  Prompt for Segmenting Camouflaged Objects},
  booktitle    = AAAI,
  pages        = {12511--12518},
  year         = {2024}
}

@inproceedings{Zhang2024ACUMEN,
  author       = {Hong Zhang and
                  Yixuan Lyu and
                  Qian Yu and
                  Hanyang Liu and
                  Huimin Ma and
                  Ding Yuan and
                  Yifan Yang},
  title        = {Unlocking Attributes' Contribution to Successful Camouflage:
                  {A} Combined Textual and Visual Analysis Strategy},
  booktitle    = ECCV,
  series       = {Lecture Notes in Computer Science},
  volume       = {15113},
  pages        = {315--331},
  year         = {2024}
}

@article{Fan2020InfNet,
  author       = {Deng{-}Ping Fan and
                  Tao Zhou and
                  Ge{-}Peng Ji and
                  Yi Zhou and
                  Geng Chen and
                  Huazhu Fu and
                  Jianbing Shen and
                  Ling Shao},
  title        = {Inf-Net: Automatic {COVID-19} Lung Infection Segmentation From {CT}
                  Images},
  journal      = {{IEEE} Trans. Medical Imaging},
  volume       = {39},
  number       = {8},
  pages        = {2626--2637},
  year         = {2020}
}

@article{Liu2023MHNet,
  author       = {Maozhen Liu and
                  Xiaoguang Di},
  title        = {Extraordinary MHNet: Military high-level camouflage object detection
                  network and dataset},
  journal      = {Neurocomputing},
  volume       = {549},
  pages        = {126466},
  year         = {2023}
}

@article{Van2018InfoNCE,
  author       = {A{\"{a}}ron van den Oord and
                  Yazhe Li and
                  Oriol Vinyals},
  title        = {Representation Learning with Contrastive Predictive Coding},
  journal      = {CoRR},
  volume       = {abs/1807.03748},
  year         = {2018}
}

@article{Chen2015cococaption,
  author       = {Xinlei Chen and
                  Hao Fang and
                  Tsung{-}Yi Lin and
                  Ramakrishna Vedantam and
                  Saurabh Gupta and
                  Piotr Doll{\'{a}}r and
                  C. Lawrence Zitnick},
  title        = {Microsoft {COCO} Captions: Data Collection and Evaluation Server},
  journal      = {CoRR},
  volume       = {abs/1504.00325},
  year         = {2015},
  eprinttype    = {arXiv},
  eprint       = {1504.00325}
}

@inproceedings{Ruan2025MMCamObj,
  author       = {Jiacheng Ruan and
                  Wenzhen Yuan and
                  Zehao Lin and
                  Ning Liao and
                  Zhiyu Li and
                  Feiyu Xiong and
                  Ting Liu and
                  Yuzhuo Fu},
  title        = {MM-CamObj: {A} Comprehensive Multimodal Dataset for Camouflaged Object
                  Scenarios},
  booktitle    = AAAI,
  pages        = {6740--6748},
  year         = {2025}
}

@inproceedings{He2024TextpromptCIS,
  author       = {Zhentao He and
                  Changqun Xia and
                  Shengye Qiao and
                  Jia Li},
  title        = {Text-prompt Camouflaged Instance Segmentation with Graduated Camouflage
                  Learning},
  booktitle    = ACMMM,
  pages        = {5584--5593},
  year         = {2024}
}

@inproceedings{Luo2023CIS,
  author       = {Naisong Luo and
                  Yuwen Pan and
                  Rui Sun and
                  Tianzhu Zhang and
                  Zhiwei Xiong and
                  Feng Wu},
  title        = {Camouflaged Instance Segmentation via Explicit De-Camouflaging},
  booktitle    = CVPR,
  pages        = {17918--17927},
  year         = {2023}
}

@inproceedings{Sun2023COC,
  author       = {Guolei Sun and
                  Zhaochong An and
                  Yun Liu and
                  Ce Liu and
                  Christos Sakaridis and
                  Deng{-}Ping Fan and
                  Luc Van Gool},
  title        = {Indiscernible Object Counting in Underwater Scenes},
  booktitle    = CVPR,
  pages        = {13791--13801},
  year         = {2023}
}

@article{Zhao2025CvpAgent,
  author       = {Pancheng Zhao and
                  Deng{-}Ping Fan and
                  Shupeng Cheng and
                  Salman H. Khan and
                  Fahad Shahbaz Khan and
                  David A. Clifton and
                  Peng Xu and
                  Jufeng Yang},
  title        = {Deep Learning in Concealed Dense Prediction},
  journal      = {CoRR},
  volume       = {abs/2504.10979},
  year         = {2025}
}

@inproceedings{Pang2024OVCOS,
  author       = {Youwei Pang and
                  Xiaoqi Zhao and
                  Jiaming Zuo and
                  Lihe Zhang and
                  Huchuan Lu},
  title        = {Open-Vocabulary Camouflaged Object Segmentation},
  booktitle    = ECCV,
  volume       = {15105},
  pages        = {476--495},
  year         = {2024}
}

@inproceedings{liu2025D2SVSE,
  title={Aligning Information Capacity Between Vision and Language via Dense-to-Sparse Feature Distillation for Image-Text Matching},
  author={Liu, Yang and Feng, Wentao and Liu, Zhuoyao and Huang, Shudong and Lv, Jiancheng},
  booktitle=ICCV,
  pages  = {21679--21688},
  year={2025}
}

@inproceedings{liu2025AVSE,
  title={Asymmetric Visual Semantic Embedding Framework for Efficient Vision-Language Alignment},
  author={Liu, Yang and Liu, Mengyuan and Huang, Shudong and Lv, Jiancheng},
  booktitle=AAAI,
  volume={39},
  number={6},
  pages={5676--5684},
  year={2025}
}

@inproceedings{pan2023CHAN,
  title={Fine-grained image-text matching by cross-modal hard aligning network},
  author={Pan, Zhengxin and Wu, Fangyu and Zhang, Bailing},
  booktitle=CVPR,
  pages={19275--19284},
  year={2023}
}

@inproceedings{Dosovitskiy2021ViT,
  author       = {Alexey Dosovitskiy and
                  Lucas Beyer and
                  Alexander Kolesnikov and
                  Dirk Weissenborn and
                  Xiaohua Zhai and
                  Thomas Unterthiner and
                  Mostafa Dehghani and
                  Matthias Minderer and
                  Georg Heigold and
                  Sylvain Gelly and
                  Jakob Uszkoreit and
                  Neil Houlsby},
  title        = {An Image is Worth 16x16 Words: Transformers for Image Recognition
                  at Scale},
  booktitle    = ICLR,
  year         = {2021}
}

@inproceedings{Deng2009ImageNet21K,
  author       = {Jia Deng and
                  Wei Dong and
                  Richard Socher and
                  Li{-}Jia Li and
                  Kai Li and
                  Li Fei{-}Fei},
  title        = {ImageNet: {A} large-scale hierarchical image database},
  booktitle    = CVPR,
  pages        = {248--255},
  year         = {2009}
}

@inproceedings{HuangWW2017SMLSTM,
  author       = {Yan Huang and
                  Wei Wang and
                  Liang Wang},
  title        = {Instance-Aware Image and Sentence Matching with Selective Multimodal
                  {LSTM}},
  booktitle    = CVPR,
  pages        = {7254--7262},
  year         = {2017}
}

@inproceedings{EisenschtatW2017ttwoWayNets,
  author       = {Aviv Eisenschtat and
                  Lior Wolf},
  title        = {Linking Image and Text with 2-Way Nets},
  booktitle    = CVPR,
  pages        = {1855--1865},
  year         = {2017}
}

@inproceedings{VendrovKFU2015OrderEmbeddings,
  author       = {Ivan Vendrov and
                  Ryan Kiros and
                  Sanja Fidler and
                  Raquel Urtasun},
  title        = {Order-Embeddings of Images and Language},
  booktitle    = ICLR,
  year         = {2016}
}

@inproceedings{Luo2021BUTDFailure,
  author       = {Yunpeng Luo and
                  Jiayi Ji and
                  Xiaoshuai Sun and
                  Liujuan Cao and
                  Yongjian Wu and
                  Feiyue Huang and
                  Chia{-}Wen Lin and
                  Rongrong Ji},
  title        = {Dual-level Collaborative Transformer for Image Captioning},
  booktitle    = AAAI,
  pages        = {2286--2293},
  year         = {2021}
}

@inproceedings{Fan2017StructureMeasureAN,
  title={Structure-Measure: A New Way to Evaluate Foreground Maps},
  author={Deng-Ping Fan and Ming-Ming Cheng and Yun Liu and Tao Li and A. Borji},
  booktitle=ICCV,
  year={2017},
  pages={4558-4567}
}
}

\end{document}